\documentclass{article}

\usepackage{microtype}
\usepackage{graphicx}
\usepackage{subfigure}
\usepackage{booktabs} 
\usepackage{amssymb}
\usepackage{amsmath}
\usepackage{todonotes}
\usepackage{multirow}
\usepackage[title]{appendix}

\usepackage{hyperref}

\usepackage[accepted]{icml2020}

\icmltitlerunning{Self-Paced Absolute Learning Progress}

\begin{document}

\twocolumn[
\icmltitle{\texorpdfstring{Self-Paced Absolute Learning Progress as a Regularized Approach \\ to Curriculum Learning}{Self-Paced Absolute Learning Progress as a Regularized Approach to Curriculum Learning} }



\icmlsetsymbol{equal}{*}

\begin{icmlauthorlist}
\icmlauthor{Ulla Scheler}{equal,to}
\icmlauthor{Tobias F. Niehues}{equal,to}
\icmlauthor{Pascal Klink}{to}
\end{icmlauthorlist}

\icmlaffiliation{to}{Department of Intelligent Autonomous Systems, Technical University of Darmstadt, Darmstadt, Germany}

\icmlcorrespondingauthor{Ulla Scheler}{ulla.scheler@stud.tu-darmstadt.de}
\icmlcorrespondingauthor{Tobias F. Niehues}{niehues.tobias@gmail.com}

\icmlkeywords{Reinforcement Learning, Curriculum Learning, Intrinsic Motivation, Self-Paced Learning, ICML}

\vskip 0.3in
]



\printAffiliationsAndNotice{*Equal Contribution} 

\begin{abstract}

The usability of Reinforcement Learning is restricted by the large computation times it requires. Curriculum Reinforcement Learning speeds up learning by defining a helpful order in which an agent encounters tasks, i.e. from simple to hard.
Curricula based on Absolute Learning Progress (ALP) have proven successful in different environments, but waste computation on repeating already learned behaviour in new tasks. We solve this problem by introducing a new regularization method based on Self-Paced (Deep) Learning, called Self-Paced Absolute Learning Progress (SPALP). We evaluate our method in three different environments. Our method achieves performance comparable to original ALP in all cases, and reaches it quicker than ALP in two of them. We illustrate possibilities to further improve the efficiency and performance of SPALP.

\end{abstract}

\section{Introduction}\label{introduction}
In Reinforcement Learning (RL) an agent adapts its actions to maximize a reward signal it receives from its environment, e.g. \cite{sutton2018reinforcement}. In combination with methods from Deep Learning, RL was successfully employed in virtual gaming environments like Atari \cite{mnih_playing_2013} or Starcraft \cite{vinyals_grandmaster_2019}, as well as in physical manipulation tasks, e.g. \cite{andrychowicz_learning_2020, rajeswaran_learning_2018, kalashnikov_scalable_2018}. Before these levels of performance are reached, though, an agent often takes millions of steps while it tries different actions (this is called 'exploration') and encounters different variants of the environment. Instead of optimizing the agent's exploration of the parameter space, Curriculum Learning (CL) \cite{bengio2009, matiisen2020} tries to systematically present the agent with an optimal order of the parameter space samples. In this context, the algorithm that chooses the next training environment based on distributions generated by itself is often referred to as the teacher, while the executing agent is correspondingly called student.\\
There are several approaches to implementing a curriculum, for example deriving goal tasks via GANs \cite{florensa2018}, incrementally adjusting the sampling distribution to resemble the distribution of a goal task, known as Self-Paced Deep Reinforcement Learning (SPDL) \cite{klink2020b}, and sampling techniques that sample the parameter spaced based on recent learning advancement as in Absolute Learning Progress (ALP) \cite{portelas2020}.\\
In this work, we apply insights from Self-Paced Learning as regularization in ALP settings and evaluate this approach in various environments.

\subsection{Self-Paced Learning}\label{spl}

The usual goal of self-paced learning \cite{klink2020b, klink2020a} is to master a goal task, defined as a subspace of the full parameter space, by gradually adjusting a distribution of tasks  $p(c|\nu)$ used for sampling the training contexts $c$ and its parameters $\nu$ to the goal task distribution $\mu(c)$. This can be accomplished by incorporating the KL-divergence between the sampling and the goal distribution into the objective function of the expected reward \eqref{eq:spl_objective}, given the reward function $R(\omega, c)$ and the policy $\pi_\omega$, parameterized by $\omega$. The parameter $\alpha$ controls the weight of the KL-divergence relative to the reward objective.

\begin{equation}\label{eq:spl_objective}
    \max_{\omega, \nu} \mathbb{E}_{p(c|\nu)} [R(\omega, c)] - \alpha D_{KL}(p(c|\nu) \parallel \mu(c))
\end{equation}

Equation \eqref{eq:spl_objective} can be solved by alternately maximizing w.r.t. $\omega$ and $\nu$ while gradually increasing $\alpha$ to converge to the goal distribution. In other words, the teacher attempts for the student to receive a good reward, while also exposing it to ever closer approximations of the goal environment.\\
Recent work in self-paced learning \cite{klink2021} has found that optimizing this modified objective function is equal to maximizing objective \eqref{eq:spl_regularized_objective} w.r.t the sampling parameters $\omega$. \\
Please note that this version was originally derived from a supervised learning scenario in which the objective function aimed to minimize losses \cite{kumar2010, jiang2015}. Since application to an RL setting involves transforming the objective function from a loss minimization to a reward maximization and all losses previously were required to be all positive, the rewards fed into \eqref{eq:spl_regularized_objective} now need to be all negative, e.g. by an appropriate transformation beforehand.

\begin{equation}\label{eq:spl_regularized_objective}
    \max_{\omega} \mathbb{E}_{\mu(c)} [-\alpha (1 - \exp(\frac{1}{\alpha}R(\omega, c)))]
\end{equation}

Note that SPDL only indirectly includes information about past experiences and learning progress in form of the learner's policy and the teacher's sampling distribution: The learner prefers to repeat behavior that previously performed well while the teacher fits its sampling distribution taking into account the rewards from previous iterations. In our approach we will combine SPDL's regularization objective with a more explicit take on learning progress and past experience.

\subsection{Absolute Learning Progress (ALP)}\label{alp}
The Absolute Learning Progress model (ALP-GMM) proposed by \citet{portelas2020} is an approach to model intrinsic motivation in reinforcement learning \cite{oudeyer2007, baranes2013}, which is inspired by models of intrinsic motivation in humans \cite{moulin-frier2014}. The term 'GMM' refers to the use of Gaussian Mixture models used to represent the sampling distribution.\\
ALP-GMM uses the so-called Absolute Learning Progress as a measure of how much progress was gained by training on the current task \eqref{eq:alp}. Tasks and their ALPs are saved into a reward history. The reward history is implemented as a first-in-first-out buffer of fixed size. To calculate the ALP for a reward $r_\text{new}$ for a task $\theta_\text{new}$ we look up the nearest neighbor $\theta_\text{NN}$ in the reward history and use its reward $r_\text{NN}$:

\begin{equation}\label{eq:alp}
    \text{ALP} = |r_\text{new} - r_\text{NN}|
\end{equation}

\begin{figure*}[tb]
\vskip 0.2in
\begin{center}
\centerline{\includegraphics[width=\linewidth]{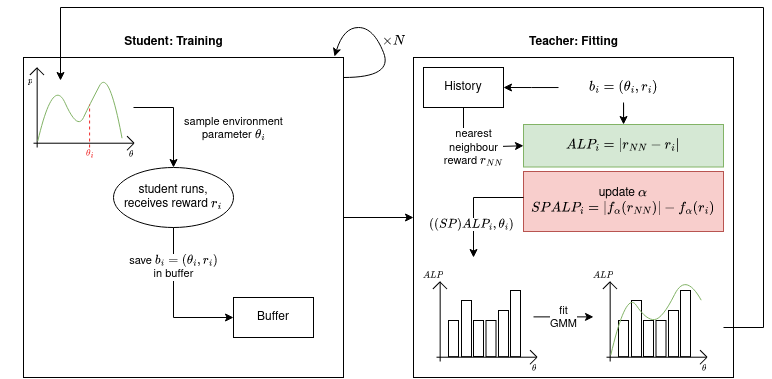}}
\caption{\textbf{A high-level illustration of how ALP-GMM and SPALP work}. Note that the only difference between the two algorithms lies in how the ALP are calculated. Left side: The student makes N runs. Environments are sampled from current teacher distribution each run. Right side: Calculate ALP for each parameter-reward-pair $b_i$ in Buffer and fit GMM.}
\label{fig:alp:illustration}
\end{center}
\vskip -0.2in
\end{figure*}

The sampling distribution during training is defined by a Gaussian mixture model (GMM) fitted to the ALPs of the batch of the last $N$ sampled tasks. The next task is then with a probability $p$ either sampled from the GMM or uniformly from the parameter space, to ensure exploration. A basic overview of ALP-GMM can be found in Figure \ref{fig:alp:illustration}. By basing most of the sampling on ALPs, the agent is provided with samples from subspaces of the parameter space that yielded the most progress in the last iterations. At the same time, the algorithm remembers the progress from older examples in form of the reward history. 

One of the biggest advantages of ALP-GMM over other Curriculum Learning approaches is its quite simple concept, for it does not need any structures like Markov Decision Processes (MDPs) or multi-armed bandits, as used by \citet{matiisen2020} or \citet{graves2017}. \\
Furthermore, it only needs tuning of relatively few hyperparameters. One must basically only provide hyperparameters defining how to fit the GMM, as the ALP metric itself does not need any additional hyperparameters.

Yet, please note that ALP-GMM does not differentiate whether the progress was achieved in a task where the agent is already quite proficient or in one where its performance was previously low. It cares only about the absolute difference between the current performance and the closest previous performance saved in the reward history. Thus ALP-GMM might waste computation on repeating already learned behaviour in new tasks or tasks it has not seen in a while. In these cases the ALP might be high, but does not signify any real learning progress.\\
The regularizer we introduce will extend ALP by mostly preferring environment parameterizations which yield higher rewards. The logic behind this is the following: Progress in a task from a lower to a low reward might not mean progress in the policy, i.e. the student most likely only utilizes what it already learned somewhere else but is not improving its overall behavior. In contrast, progress in a task from a high reward $r_i$ to a higher reward $r_{i+1}$ is more likely to stem from actual new learning as simple behaviors were probably already used to reach $r_i$ and will not lead to further improvements.

\begin{figure}[!hbtp]
\vskip 0.2in
\begin{center}
\centerline{\includegraphics[width=\linewidth]{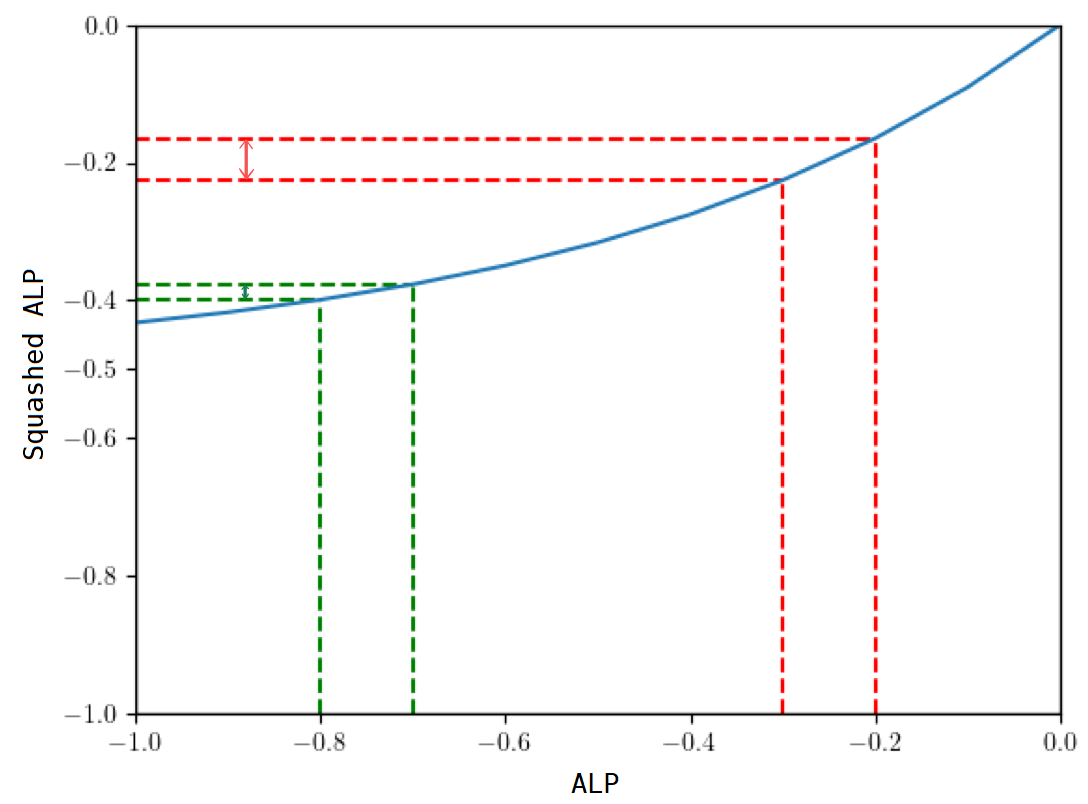}}
\caption{\textbf{Effect of squashing function $f_\alpha(x)$ on ALP calculation in SPALP}: Higher rewards are squashed less and thus the same absolute difference in rewards leads to a larger overall ALP. Squashing function plotted with $\alpha = 0.5$.}
\label{fig:alp:squashing_vs_nonsquashing}
\end{center}
\vskip -0.2in
\end{figure}

\newpage
\section{Self-Paced Absolute Learning Progress}\label{spalp}

We introduce Self-Paced Absolute Learning Progress (SPALP) as a variant of the ALP-GMM model, making use of the regularization in the modified objective function \eqref{eq:spl_regularized_objective} from SPDL. This yields a regularization function for the rewards used when calculating the absolute learning progress given by \eqref{eq:spalp:regularization}, and leads to our so-called regularized ALP \eqref{eq:spalp:regularized_alp}. An overview of our approach can be found in Algorithm \ref{alg:spalp} and Figure \ref{fig:alp:illustration}.\\

\begin{equation}\label{eq:spalp:regularized_alp}
    \text{ALP} = |f_\alpha(r_\text{new}) - f_\alpha(r_\text{old})|
\end{equation}

We modify all rewards received from the different environments in two ways: Firstly, as ranges of rewards differ between environments, we normalize all environments to the interval $[0,1]$. Secondly, since the regularization requires all rewards to be negative (as mentioned in \ref{spl}), we shift these normalized rewards by $-1$ into the interval $I_r = [-1; 0]$ before applying the regularization.

\begin{equation}\label{eq:spalp:regularization}
    f_\alpha(x) = -\alpha (1 - \exp(\frac{x}{\alpha}))
\end{equation}

\hfill \break

\begin{figure}[tb]
\vskip 0.2in
\begin{center}
\centerline{\includegraphics[width=\columnwidth]{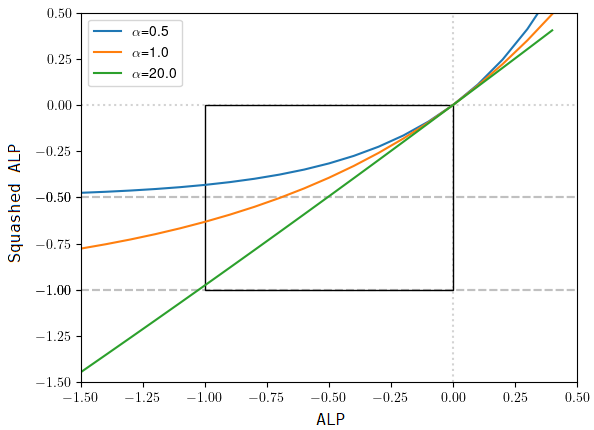}}
\caption{\textbf{Squashing function $f_\alpha(x)$} for different values of alpha. In negative direction each curve converges to the negative $\alpha$-values (indicated by the dashed lines), while they increase exponentially in the positive direction. Since all rewards are normalized into $[-1.0; 0.0]$, this is the interval of actual relevance for our work. Higher values of $\alpha$ lead to a more linear behavior, while smaller values lead to a stronger squashing of smaller rewards.}
\label{fig:alpha_squashing}
\end{center}
\vskip -0.2in
\end{figure}

The function $f_\alpha(x)$, shown in Figure \ref{fig:alpha_squashing}, can be thought of as a softener on ALPs calculated from low rewards. An example of this can be seen in Figure \ref{fig:alp:squashing_vs_nonsquashing}. Smaller values of $\alpha$ lead to more squashing. Rewards are squashed into the interval $[-\alpha, 0.0]$ instead of the regular interval $I_r$. Low rewards are squashed more strongly than high rewards. Thereby the ALPs calculated from low squashed rewards also become smaller. High rewards remain mostly untouched, because $f_\alpha$ approaches linear behavior near the origin. Thus, the regularization employs a rather conservative sampling strategy by focusing on tasks that already yield high rewards.\\
For high values of $\alpha$ the regularization function $f_\alpha$ converges to the identity mapping in $I_r$ and we can think of the regularization as being turned off in this case.

For determining the exact value of $\alpha$ and thus the strength of the squashing in the regularization function $f_\alpha$, we needed an additional condition. Our solution is to choose $\alpha$ such that it fulfills equation \eqref{eq:spalp:alpha_condition}. That is, we choose $\alpha$ in a way that the squashed rewards are reaching a bound $r_b$, which is a new hyperparameter specific to SPALP. The bound $r_b$ can intuitively be understood as the mean reward we are striving for in the region of the parameter space previously sampled from. If the rewards $r_i$ were low, we need a stronger regularization to reach $r_b$, resulting in a low value of $\alpha$ and vice versa.

\begin{equation}\label{eq:spalp:alpha_condition}
   \frac{1}{N}\sum_if_{\alpha}(r_i) = r_b
\end{equation}

By choosing $\alpha$ the way we do, the regularization gets weaker the closer the mean reward gets to the bound $r_b$, i.e. in general the longer the agent trains, the weaker the regularization becomes. To further support this behavior, we turn the regularization off completely if the mean reward of the last $N$ samples exceeded $r_b$ and fall back to vanilla ALP-GMM. This ensures exploration, since the regularization introduces a bias by focusing on tasks on which we were already performing well, while the unregularized version treats all ALPs equally (cf. figure \ref{fig:alp:squashing_vs_nonsquashing}). Once the unregularized mean reward falls below $r_b$, we turn it back on again.


The complete implementation is shown by Algorithm \ref{alg:spalp}. Please note that we are not only sampling from the fitted GMM, but also uniformly take random samples from the parameter space with a probability of $p=0.2$ to ensure exploration independent of the GMM, just as vanilla ALP-GMM does.

\begin{algorithm}[htbp]
   \caption{Self-Paced Absolute Learning Progress}
   \label{alg:spalp}
\begin{algorithmic}[1]
   \STATE {\bfseries Input:} fitting rate $N$, bounded parameter space $\mathcal{P}$, student $\mathcal{S}$, teacher $\mathcal{T}$
   \STATE {\bfseries Initialize} First-in-First-Out buffer
   $\mathcal{B}$ with max size $N$
   \STATE {\bfseries Initialize} reward history database $\mathcal{H}$
   
   \STATE {\bfseries Bootstrapping} for $N$ iterations
   \WHILE{not terminated}
   \STATE update $\alpha$ 
   \STATE calculate ALP of last $N$ steps using the reward of the nearest neighbor (in parameter space) in $H$
   \STATE fit teacher-GMM on ALPs 
   
    \FOR{$N$ runs}
   \STATE sample $\theta_i$ from $\mathcal{P}$ using the GMM of $\mathcal{T}$
   \STATE write reward $r_i$ from $\mathcal{S}$ in environment $E_{\theta}$ into $B$
    \ENDFOR
   
   \ENDWHILE
   \STATE {\bfseries return} $\mathcal{S}$
\end{algorithmic}
\end{algorithm}




\section{Experiments}\label{experiments}

For evaluation and comparison of our teaching algorithm we used three different environments as shown in Appendix \ref{appendix:environments_overview}; a toy environment where the progress of rewards in the sample space is approximated by a mathematical function for different subspaces, the bipedal walker environment as used by \citet{portelas2020} for evaluation of ALP-GMM and the ball catching environment as used by \citet{klink2020b, klink2021} for evaluation of their Self-Paced Deep Reinforcement Learning.

\subsection{Toy Environment}\label{experiments:toy_env}
The toy environment already provided by \citet{portelas2020} subdivides an $n$-dimensional space into (hyper)cubes along each dimension. To simulate in making progress on the rewards, sampling from a cube linearly increases the reward for sampling from this very cube the next time. At first, only the cube in the bottom left (equivalent to the supposedly easiest parameter configurations) yields a reward when being sampled from. All other cubes need to be unlocked first by having a `mastered' hypercube in their immediate vicinity. A cube is `mastered' if it yielded a reward higher than 75. The maximum reward for each cube is capped at 100. \\
We extended this environment by incorporating additional reward progress shapes besides the aforementioned linear behavior. To achieve a more realistic change in the rewards, we found a sigmoidal curve to be an adequate choice, as the reward needs some time at first to increase more strongly (equivalent to an agent getting used to a task for the first time) and only yields small progress before saturating to the reward maximum (equivalent to the agent possibly not being able to make any more progress). \\
Further, we wanted to also add a concept of transfer learning, where newly discovered cubes benefit from performances in neighboring cubes in their direct vicinity. For this, when sampling from each cube for the first time, it already yields a reward equal to the average reward already obtained from its neighboring cubes.

Our main metric used for evaluation of different algorithms in this environment was the percentage of `mastered' hypercubes over the whole parameter space. Table \ref{tab:toy_env_results} shows this performance in comparison to random sampling and ALP-GMM. \\
$r_b=-0.4$ resulted in the best performance remarkably often; still, this is not in line with the results of the other environments, shown in sections \ref{experiments:bipedal_walker_env} and \ref{experiments:ball_catching_env}.

\begin{table}[th]
\caption{\textbf{Performance in the toy environment for different configurations} ($\pm$ standard error). `TL' indicates whether Transfer Learning was simulated. Performance measured as percentage of mastered hypercubes. For $d=2$, each dimension consisted of 10 cubes; 5 cubes per dimension for $d=3$. 3 runs per configuration with 40000 episodes each. For SPALP the best performing $r_b$ was chosen each time. Performances of the best teachers per configuration are highlighted bold.}
\label{tab:toy_env_results}
\vskip 0.15in
\begin{center}
\begin{small}
\begin{sc}
\begin{tabular}{c|ccccr}
\toprule
$d$ & reward & tl & teacher & $r_b$ &  performance \\
\midrule
  \multirow{12}{*}{2}&       \multirow{6}{*}{linear}     &\multirow{3}{*}{$\times$} &      random &       - &        14.67 $\pm$             0.17 \\
 &  &  &      alpgmm &       - &        49.89 $\pm$             1.47 \\

  &            & &  spalp &       -0.3 &        \textbf{53.00 $\pm$             7.51} \\
  \cline{3-6}
    &            &  \multirow{3}{*}{$\surd$}&      random &       - &        30.70 $\pm$             0.37 \\
  &            &   &      alpgmm &       - &        80.86 $\pm$             2.29 \\

  &            &  &  spalp &       -0.4 &        \textbf{91.33 $\pm$             2.19} \\
\cline{2-6}
  &      \multirow{6}{*}{sigmoid}      & \multirow{3}{*}{$\times$}&      random &       - &        26.96 $\pm$             0.29 \\
  &     &  &      alpgmm &       - &        43.41 $\pm$             1.62 \\
  &            & &  spalp &       -0.4 &        \textbf{52.33 $\pm$            6.69} \\
  \cline{3-6}
  &            &  \multirow{3}{*}{$\surd$}  &      random &       - &        58.63 $\pm$             0.57 \\
  &            & &      alpgmm &       - &        85.26 $\pm$             1.47 \\
  
  &            &  &  spalp &       -0.5 &        \textbf{91.00 $\pm$             3.51} \\
\hline
 \hline
  \multirow{12}{*}{3} &     \multirow{6}{*}{linear}       & \multirow{3}{*}{$\times$}&      random &       - &        15.47 $\pm$             0.15 \\
 &  &  &      alpgmm &       - &        47.79 $\pm$             2.98 \\

  &            & &  spalp &       -0.4 &        \textbf{65.07 $\pm$             1.48} \\
  \cline{3-6}
    &            & \multirow{3}{*}{$\surd$}   &      random &       - &        57.69 $\pm$             1.51 \\
  &            &  &      alpgmm &       - &        96.95 $\pm$             1.56 \\

  &            &  &  spalp &       -0.4 &      \textbf{ 100.00 $\pm$             0.00} \\
\cline{2-6}
  &         \multirow{6}{*}{sigmoid}   & \multirow{3}{*}{$\times$}&      random &       - &        33.24 $\pm$             0.79 \\
  &     &  &      alpgmm &       - &        42.79 $\pm$             1.87 \\

  &            & &  spalp &       -0.9 &        \textbf{52.00 $\pm$             7.43} \\
  \cline{3-6}
    &            &  \multirow{3}{*}{$\surd$} &      random &       - &        99.82 $\pm$             0.08 \\
  &            &   &      alpgmm &       - &        92.83 $\pm$             2.17 \\

  &            &  &  spalp &       -0.6 &       \textbf{100.00 $\pm$             0.00} \\
\bottomrule
\end{tabular}

\end{sc}
\end{small}
\end{center}
\vskip -0.1in
\end{table}

\subsection{Bipedal Walker Environment}\label{experiments:bipedal_walker_env}
We use the Bipedal Walker environment by \citet{portelas2020} in the standard configuration provided by the author (default leg length of the walker, obstacle spacing from 0 to 6 and obstacle height from 0 to 3), which is in turn adapted from \citet{openai_bipedal_walker}.
\\
The goal of the agent in this environment is to walk as long as possible without falling down. Reward is received for moving forward, while the application of motor torque costs a small amount of points and falling down is penalized and ends the episode. Falling is caused either by the walker's inability to walk or inability to cross an obstacle. Note that the parameter space, consisting of obstacle spacing and obstacle height, contains regions that are infeasible due to the agent's leg length. The state of the agent consists of several velocity measures of the joints, 10 lidar rangefinder measurements and legs contact with the ground.\\

As student we used SAC \cite{haarnoja2018} from the \texttt{Stable Baselines} library \cite{stable-baselines}, again slightly altered by \cite{portelas2020}. For comparison, ALP-GMM was always run with the parameters provided by \citet{portelas2020}.\\
We evaluated the performance of the teachers by evaluating how many of 50 different walker environments they were able to master. Because these test environments were chosen at random from all over the parameter space, i.e. with varying obstacle spacing and obstacle height, there are some that were infeasible to the default walker due to its leg length. An environment counts as "mastered" if the student reached more than 230 points. This arbitrary bound was again modeled after \cite{portelas2020}.\\

\begin{figure}[tb]
\vskip 0.2in
\begin{center}
\centerline{\includegraphics[width=\columnwidth]{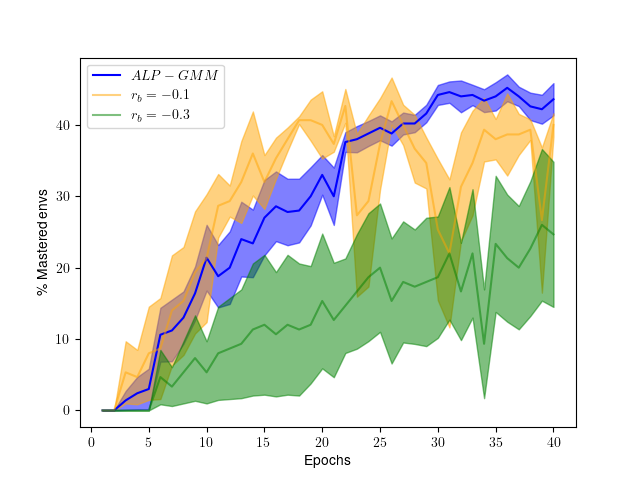}}
\caption{\textbf{Percentage of mastered environments for the bipedal walker environment}. After each training epoch the student is tested in 50 environments of the parameter space. An environment counts as "mastered" if the student achieves more than a certain reward. Figure shows mean performance and standard error of 10 (ALP-GMM) or 3 (SPALP) seeds.}
\label{fig:walker_bound}
\end{center}
\vskip -0.2in
\end{figure}

The results are shown in figure \ref{fig:walker_bound}. Please note that due to time and computing constraints, the SPALP results represent only three seeds and correspondingly more noise. To illustrate: The dips in the SPALP curves are due to noise of an errant seed. SPALP with $r_b=-0.1$ reaches an overall performance similar to ALP-GMM but does so earlier. In the case of $r_b=-0.1$, the reward bound was not reached in any of the seeds and correspondingly, squashing was always turned on. To show the negative influence of a badly chosen reward bound, we also included the performance of the worst bound ($r_b=-0.3$). We should note that low $r_b$ ($-0.8; -0.9$) also performed well in the bipedal walker environment. In these cases the reward bound was reached before the first fitting of the teacher distribution, though. As this is equivalent to running vanilla ALP-GMM, we did not include these runs in the results.\\

\begin{figure}[!b]
\begin{center}
\centerline{\includegraphics[width=\columnwidth]{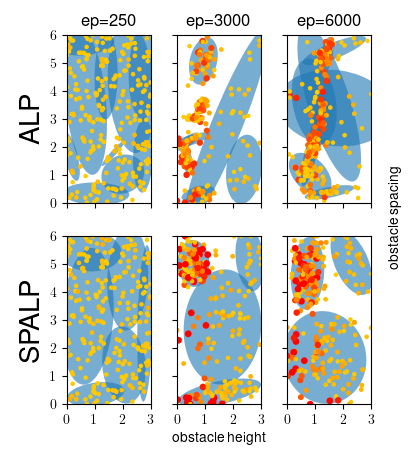}}
\caption{\textbf{Different sampling behaviors of ALP-GMM and SPALP}, illustrated in the bipedal walker environment for episodes 250, 3000 and 6000. Blue ellipses represent the GMMs, the dots represent single samples. Yellow samples represent a low (regularized) ALP, red samples represent a high (regularized) ALP. Both algorithms start with a general exploration period of 250 episodes. After that, SPALP focuses its sampling on the easiest subspace of the parameter space (low and widely spaced obstacles, upper left corner). In contrast, ALP-GMM samples more broadly in the low obstacle height part of the parameter space. Although the data represents a single, successful seed per algorithm of ALP-GMM and SPALP with $r_b=-0.1$, the patterns were consistent over all successful seeds.}
\label{fig:gmms}
\end{center}
\vskip -0.5in
\end{figure}

To study the differences in the teacher's sampling distributions, we plotted the Gaussian Mixture Models after each fitting period. Across all the runs we analyzed, it became apparent that SPALP focuses on the easiest subspace of the parameter space, while ALP advances to more difficult regions more quickly (moving the concentration of samples diagonally to the bottom right in the parameter space). This is shown exemplarily for two successful seeds in figure \ref{fig:gmms}. In the case of unsuccessful SPALP seeds, we noticed that from the beginning the teacher did not focus on the easiest subspace to the same extent.

\subsection{Ball Catching Environment}\label{experiments:ball_catching_env}
The ball catching environment is the same as used by \citet{klink2020b, klink2021}; consult his paper for any experimental details on the simulation please, since for our purposes not all simulation details are relevant\footnote{For simulation the \href{http://www.mujoco.org}{MuJoCo physics engine} was used.}.\\
The environment is defined by the distance from which the ball is thrown and its target position in the catching plane of the robot. The reward function of this task is sparse, as the agent only receives a reward when catching the ball but control cost for any movements made at all times. \\
To provide the robot with a good initial behavior and the possibility to focus on the catching task instead of learning its plain controls, it was pre-trained to a policy by which it is capable of holding its initial position.

\begin{figure}[tb]
\vskip 0.2in
\begin{center}
\centerline{\includegraphics[width=\columnwidth]{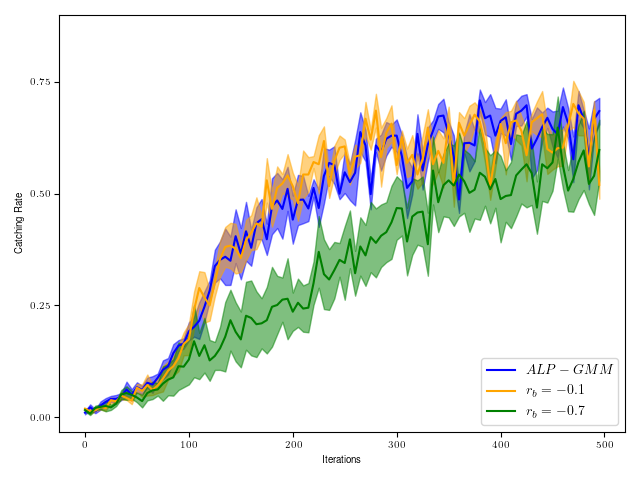}}
\caption{\textbf{Percentage of successful catches of 200 throws in the ball catching environment} after training with ALP-GMM and SPALP (different $r_b$). The figure shows mean performance and standard error for five seeds per parameter value.}
\label{fig:ball_catching_bound}
\end{center}
\vskip -0.2in
\end{figure}

For training we used SAC \cite{haarnoja2018} as the student, as provided by the \texttt{Stable Baselines} library \cite{stable-baselines} and used by \citet{klink2020b, klink2021}. The GMMs in ALP and SPALP were fitted after each 200 iterations during training. \\
We then evaluated the performance of the trained model for different values of $r_b$ in SPALP by running 200 runs on random ball catching tasks and measuring the percentage of successful catches, exemplary shown in Figure \ref{fig:ball_catching_bound}. The value of $r_b=-0.1$ was one of the best performers while $r_b=-0.7$ was one of the worst. In general, high $r_b$ ($-0.1; -0.2$) and low $r_b$ ($-0.8; -0.9$) performed well in the ball catching environment. Yet, for low $r_b$ the bound was often reached almost instantly, so the regularization was never put to work in the first place - for this reason we excluded those results.\\
In contrast to the bipedal walker environment (see section \ref{experiments:bipedal_walker_env}), we did not find any prominent differences in the ALP and SPALP teachers' sampling behaviors for the ball catching environment. Samples yielding a high ALP were spread over a large section of the parameter space.

\begin{figure}[tb]
\vskip 0.2in
\begin{center}
\centerline{\includegraphics[width=\columnwidth]{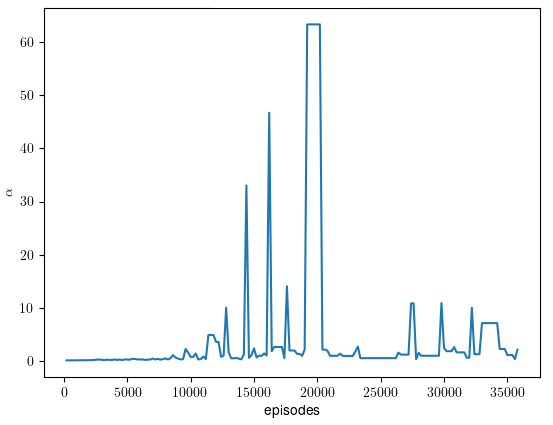}}
\caption{\textbf{Evolution of $\alpha$} for one sample run in the ball catching environment with hyperparameter $r_b=-0.1$. Most of the time the value stays below 1, but is spiking to really high values every now and then. These spikes can be explained by choosing a new $\alpha$ when the mean reward is already close to $r_b$. Since we choose $\alpha$ s. t. the mean regularized reward is equal to $r_b$ - cf. equation \eqref{eq:spalp:alpha_condition} -, this leads to a high $\alpha$ and thus a weak regularization. This is also supported by spikes regularly occuring before plateaus. The plateaus indicate every time the bound $r_b$ was surpassed by the mean reward and we fall back to vanilla ALP-GMM, thereby also turning off any $\alpha$-updates.}
\label{fig:alpha_evolution}
\end{center}
\vskip -0.2in
\end{figure}

We also evaluated how the value of $\alpha$ changed during training runs. Low values of $\alpha$ lead to strong squashing, which is only needed in order to fulfill equation \eqref{eq:spalp:alpha_condition} when the mean reward is far from the bound $r_b$. Because of this we supposed that high values of $\alpha$ (and thus a weak regularization) should occur shortly before surpassing $r_b$; cf. figure \ref{fig:alpha_squashing} for more intuition on that. Our findings support that hypothesis, as figure \ref{fig:alpha_evolution} shows.

\section{Discussion}\label{discussion}

\subsection{Overall performance of SPALP}

The results from the experiments show that SPALP with a properly chosen reward bound can reach at least the same performance levels as vanilla ALP-GMM but might do so sooner. This is exactly what we would expect from a reward regularizer: SPALP uses the same semantic structures  as ALP-GMM, so it seems unreasonable to expect it to surpass the performance of ALP-GMM in absolute terms. \\
We generally expect SPALP to perform even better, i.e. quicker, than we were able to show so far. This is mostly due to two facts: First, we only performed little tuning on the parameters. Except for the reward bound, all parameters were chosen as in \cite{portelas2020}. We expect tuning to be especially beneficial for the size of the reward history and the length of the initial exploration period and explain this reasoning below. Second, SPALP performed best in the walker and the ball catching environment with a reward bound $r_b$ of $-0.1$. This value lies on the edge of the parameter space we tried during tuning, while the general tendency of the tuning was that higher bounds performed better. Performance will thus probably improve further with a higher reward bound, e.g. $r_b = -0.05$ or even $r_b=0.1$.

\subsection{A more nuanced approach to squashing}

Picking one reward bound as the only hyperparameter for how much we squash might be too crude a heuristic. The advantage of SPALP comes from focusing on an easier subspace first, and squashing is strongest the further away the student is from the reward bound. As it gets closer during training, the squashing gets weaker, allowing the student to explore environments where the overall reward is smaller, while still taking into account the learning progress, i.e. the relative reward differences.\\
In the current implementation of SPALP, as soon as the student falls below the reward bound, squashing is turned on again and the teacher again favors parameterizations of the environment that yield high rewards. If this happens to be the easiest part of the subspace where we already sampled, the agent cannot learn anything new. In this case, we enter a cycle in which the student trains on easy environments until the reward bound gets turned off, then explores for a short while until it has to train on the easy environments again. This turning-off-and-on-again pattern is indeed what we saw in case of the ball catching environment (see figure \ref{fig:alpha_evolution}). This behavior is problematic: if progress can no longer be made in the easy subspace, but only in parts of the parameter space where the tasks are harder, learning is slower and rewards are thus smaller for a longer time.\\
One solution might consist of lowering the reward bound each time it is reached, like a ladder, thereby lowering squashing overall and lessening the sampling disadvantage of lower reward subspaces.\\
Note that whether we encounter the on-off-problem is also influenced by the size of the reward history, upon which the ALP are calculated (see section \ref{spalp}): If the reward history is long enough, the teacher can remember that the student did not make sizable learning progress in the easy subspace before squashing was turned off. In this case, even if squashing is turned on again, the teacher still favors the more difficult subspaces, because squashing can only make a small ALP of a high-reward environment bigger to some extent. \\
It follows that another solution might be to change the implementation of the reward history. We could simply make the FIFO buffer bigger. Or we could change it from one single FIFO buffer for the complete parameter space to several smaller buffers that are designated for distinct parts of the parameter space. In the latter case, the teacher remembers how much progress was made in one area until we sample there again. The last solution would be more principled but would also complicate the implementation.\\
Future work might try these more nuanced approaches on its own or in combination.

\subsection{Stronger squashing might require longer initial exploration}
Any advantage of the SPALP teacher over the ALP teacher is rooted in a different distribution from which it draws the samples it presents to the student. To study this, we explored the sampling behavior of the SPALP teacher in the bipedal walker environment in section \ref{experiments:bipedal_walker_env}. 
We found that in badly performing seeds there was a pattern that the teacher did not focus as strongly on the easy subspaces of the parameter space. Typically, this happened when the very first fitting of the GMM did not see a lot of samples in this region. While this is only one reason why a seed might perform badly, it is within our control to change it and thus deserving of further analysis.\\
When squashing leads to higher ALP on high-reward tasks, the GMMs concentrate most of their probability mass on a few tasks. If the initial exploration phase is too short and the teacher did not sample from the easy regions, the GMMs are fitted to the wrong tasks. In this case, the teacher can only recover through the small amounts of samples dedicated to random exploration. Thus, we should think of high squashing as something we do only after we have some confidence that our previous samples overall came from the right spot in the parameter space. One simple way to make this more probable is to increase the period of initial exploration before the first GMM is fitted.

\subsection{Differences in the SPALP performance over environments}

In the case of the walker environment, we found differences in how the ALP-GMM and SPALP teachers sample: In well performing seeds, SPALP focused on the easiest part of the parameter space. In the case of the ball catching environment, we did not find any differences.
This goes hand in hand with the overall performance: In the walker environment, SPALP was able to learn quicker than ALP-GMM. In the ball catching environment, the results of SPALP and ALP-GMM were very similar to each other.\\
Regarding the ball catching environment, we further remarked that the samples yielding high ALP were mostly widespread over the parameter space of the ball catching environment. In this case, there is no region SPALP can focus on.\\
It bears repeating: Any advantage of the SPALP teacher over the ALP-GMM teacher is rooted in a different distribution from which it draws the environment parameterizations it presents to the student. If there is no region to focus on, we cannot expect a difference between ALP-GMM and SPALP.
Consequently, we would expect no improvement in the performance of SPALP when using a higher reward bound: if there is no clear region where squashed ALPs are higher, more squashing won't help.\\
The lack of an apparent "easy" region might also explain the success of \citet{klink2020b} over ALP-GMM in the ball catching environment, because in their approach they explicitly define a starting distribution over the parameter space for the teacher to focus on, which they then slowly expand.
Results from the toy environment also indicate that the shape of an environment‘s reward function influence the effectiveness of SPALP. The toy environment is a handcrafted approach that might not be directly comparable to more realistic applications, like the bipedal walker and ball catching environment. Nevertheless, in this simple environment we saw significant differences in the advantage of SPALP over ALP depending on the reward function.\\
Future work could determine in which environments SPALP works better or worse than standard ALP.

\section{Conclusion}\label{conclusion}
We aimed to create an improved version (SPALP) of the already well-performing ALP-GMM from \citet{portelas2020} by extending it with a new regularization method inspired by recent progress in Self-Paced Learning. A major short-coming of our work lies in the small amount of random seeds used for some evaluations. We evaluated our approach in three environments, one of which was a handcrafted approach. Only in the handcrafted approach was SPALP able to clearly beat ALP-GMM. We explained how these results might be replicated for the other two environments by stronger squashing, a longer exploration period and a longer memory of the teacher.\\ 
Our results indicate that it depends on the environment and its reward structure how much performance benefits from regularization. Future work could examine these differences in environments in a more principled way. 

\section*{Acknowledgements}

This work originated from the master's module \textit{Integrated Project: Learning Robots} at Technical University Darmstadt and we received course credit for it. This work was not distributed and is not meant to be published publicly elsewhere.

\bibliography{rl_ip_report}
\bibliographystyle{icml2020}

\newpage

\onecolumn
\begin{appendices}
\section{Environments Overview}\label{appendix:environments_overview}
\begin{figure*}[htbp]
\addtocounter{subfigure}{-3}
\centering     
\subfigure{\includegraphics[width=0.3\linewidth]{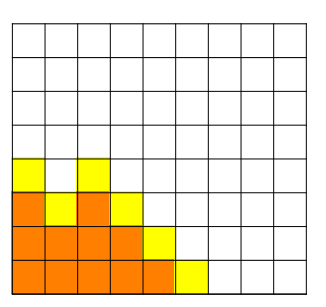}}
\hfill
\subfigure{\includegraphics[width=0.3\linewidth]{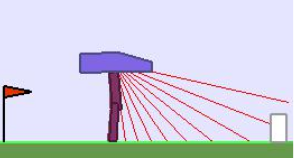}}
\hfill
\subfigure{\includegraphics[width=0.3\linewidth]{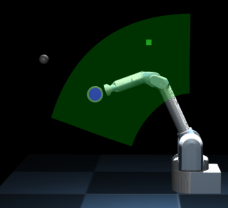}}
\smallskip
\subfigure[\textbf{Toy Environment}. Approximates a parameter space as a set of hypercubes that individually yield higher rewards proportionally to how often they were being sampled from. \newline $r_b = -0.4$ for SPALP. Originally introduced by \citet{portelas2020}.]{\label{fig:env_overview:a}\includegraphics[width=0.3\linewidth]{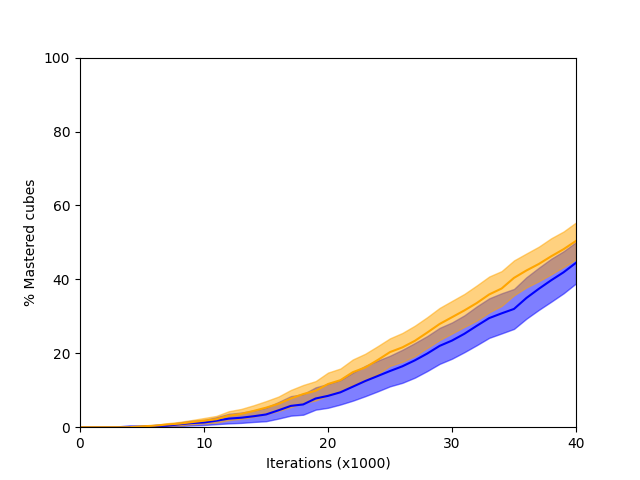}}
\hfill
\subfigure[\textbf{Bipedal Walker Environment}. The bipedal walker has to learn to walk in a way to overcome obstacles of different heights and spacings in its way. \newline $r_b = -0.1$ for SPALP. Also used for evaluation of ALP-GMM by \citet{portelas2020}. Top image taken from figure 1a of the paper by \citet{portelas2020}. ]{\label{fig:env_overview:b}\includegraphics[width=0.3\linewidth]{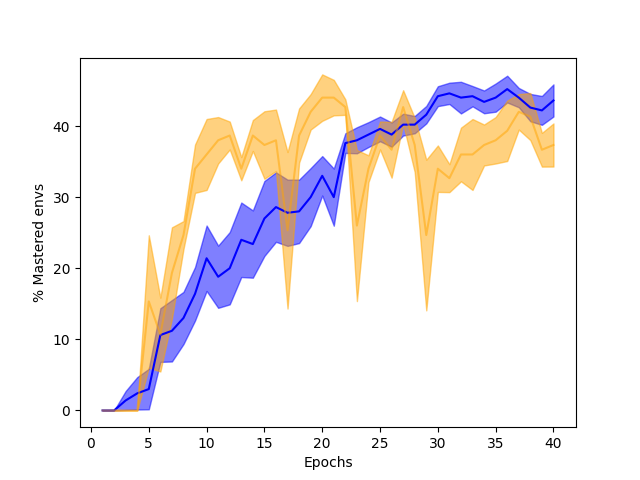}}
\hfill
\subfigure[\textbf{Ball Catching Environment}. The catcher has to learn how to move in order to catch a ball thrown somewhere in the catching plane, indicated in green in the top figure. \newline $r_b = -0.1$ for SPALP. Used by \citet{klink2020b} for evaluation of their Self-Paced Deep Learning. Top image taken from figure 1c of the paper by \citet{klink2020b}.]{\label{fig:env_overview:c}\includegraphics[width=0.3\linewidth]{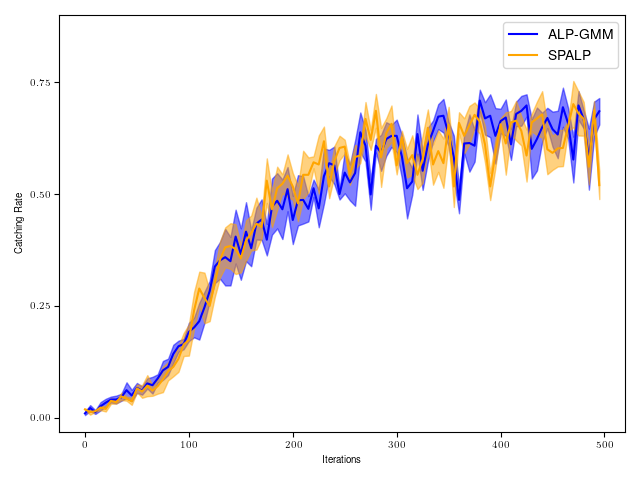}}
\caption{Renderings of the different environments and their respective best performance plots with ALP-GMM (blue) and SPALP (orange).}
\label{fig:env_overview}
\end{figure*}
\end{appendices}

\end{document}